# Mining Healthcare Procurement Data Using Text Mining and Natural Language Processing - Reflection From An Industrial Project

Ziqi Zhang[1*], Tomas Jasaitis[2], Richard Freeman[2], Rowida Alfrjani[1], Adam Funk[1]

[1] Information School, the University of Sheffield, Regent Court, Sheffield, UKS1 4DP
[firstname.lastname@sheffield.ac.uk]
[2] Vamstar Ltd., London
[firstname.lastname@vamstar.io]
(* indicates correspondence author)

**Abstract.** While text mining and NLP research has been established for decades, there remain gaps in the literature that reports the use of these techniques in building real-world applications. For example, they typically look at single and sometimes simplified tasks, and do not discuss in-depth data heterogeneity and inconsistency that is common in real-world problems or their implication on the development of their methods. Also, few prior work has focused on the healthcare domain. In this work, we describe an industry project that developed text mining and NLP solutions to mine millions of heterogeneous, multilingual procurement documents in the healthcare sector. We extract structured procurement contract data that is used to power a platform for dynamically assessing supplier risks. Our work makes unique contributions in a number of ways. First, we deal with highly heterogeneous, multilingual data and we document our approach to tackle these challenges. This is mainly based on a method that effectively uses domain knowledge and generalises to multiple text mining and NLP tasks and languages. Second, applying this method to mine millions of procurement documents, we develop the first structured procurement contract database that will help facilitate the tendering process. Second, Finally, we discuss lessons learned for practical text mining/NLP development, and make recommendations for future research and practice.

## 1. Introduction

Big data technologies have significantly impacted different industries in the last decade. The rapid increase in the amount of data has created unprecedented opportunities for cost reduction, better products and services, improved productivity and decision making. However, transforming data to knowledge that brings real 'business intelligence' remains a non-trivial, challenging task. To achieve this, data mining has been widely adopted in industries and this refers to a large range of techniques for finding patterns and correlations within data, and using such insights to predict outcomes. While data mining is often applied to structured data (e.g., database records, relational tables), it is widely acknowledged that up to 80% of 'big data' is unstructured, with text (e.g., financial report, product catalogues) representing the majority (Bach et al., 2019).

Extracting knowledge from unstructured textual data belongs to the field of text mining, which crosses the field of Natural Language Processing (NLP) that aims to make machines understand human language. These encompass a wide range of techniques such as text classification (Kowsari et al., 2019), named entity recognition (Bose et al., 2021), relation extraction (Bassignana and Plank, 2022), and terminology extraction (Dominika, 2021). And all these techniques may be used collectively to create structured knowledge bases (Krzywicki et al., 2016). Indeed, text mining and NLP research can date back to as early as the 1960s (Grishman and Sundheim, 1996) and has led to a vibrant community focusing on various tasks over the years. However, studies have shown that progress made in academic literature is not always adopted in the industrial, real-world-application context (Chiticariu et al., 2013; Krishna et al., 2016, Suganthan et al., 2015). This is often due to factors such as the difference in the evaluation focus, the timeframe for development, and the need for interpretability and after-maintenance.

For application to real-world problems, text mining finds a long history in analysing legal texts, such as for similar case matching, event timeline extraction, question and answering (Zhong et al., 2020), and judicial decision prediction (Francia et al., 2022). A significant body of work has also been done in mining Web content related to service/product provisions (Kuma et al., 2021), such as analysing social media data for customer relationship management, marketing intelligence, competitor analysis (Köseoğlu et al., 2021) and more recently, combating misinformation (e.g., fake reviews). Further, text mining has also been used in system requirements extraction and classification primarily for software engineering (Li et al., 2015; Khan et al, 2020; Tiun et al., 2020), and quality and project report analysis in the construction industry (Lee et al., 2014; Zhang et al., 2019; Tian et al., 2021). Recent studies by Rabuzin and Modrusan (2019) and Modrusan et al. (2020) identified an emergent need but a lack of application of text mining to procurement document analysis. Work in this area has only just taken off in recent years (Rabuzin and Modrusan, 2019; Modrusan et al., 2020; Choi et al., 2021; Fantoni et al., 2021; Haddadi et al., 2021).

Focusing on the application of text mining in real-world problem solving, we identify several gaps in the current literature. First, text mining for procurement is significantly under-represented but deserves increased attention from the research community. Public procurement represents a significant part of a government's financial budget and is a very complicated process involving the analysis of multiple documents at both the buyer and supplier's end. Currently, a lack of standardisation of the documentation process within and across national borders, and a lack of structured databases providing easy access to fine-grained supplier and buyer information (e.g., supplier contractual history, service/product offerings, buyer contract criteria) render the procurement process extremely time consuming and ineffective. Although some studies have looked at this area, they examined very different tasks and some lack clarity on how the end system can support real decision making (e.g., Grandia and Kruyen, 2020; Haddadi et al, 2021).

Second, we note that current studies address limited complexity in building real-world text mining applications. We describe complexity from two levels: 1) high level of heterogeneity in data, which partially leads to 2) the large range of text mining methods that need to be combined in a holistic solution. Early studies such as Chalkidis et al. (2017) investigated well-curated data sources, while work in other industrial contexts often deals with homogeneous document types ( Zhang et al., 2019; Tian et al., 2021). However, as pointed out in Modrusan et al. (2020), procurement documents are highly heterogeneous in both file format and content structure. The lack of standardisation implies a significant degree of data cleansing and in many cases, adaptation of state-of-the-art methods that are typically developed with well-curated data. This is a non-trivial process but is rarely documented in the existing literature. The complexity in the data also means that the task cannot be achieved using a single method that is often the case in the literature. Instead, a holistic approach combining multiple methods must be adopted.

Finally, a large number of studies relied on supervised methods (Chalkidis et al., 2017) that require training data and studied a single language. Training data is expensive to acquire in a business context and is often language-dependent. As a consequence, developing fully supervised methods in multilingual tasks may be infeasible for businesses. However, in many cases, businesses often possess certain forms of 'domain knowledge', such as domain specific vocabularies in Choi et al. (2021). The challenge is how to effectively use such resources across multiple tasks for many languages in a generalisable way. We noticed a lack of reporting on this particular problem.

This work fills these gaps through documenting an industry project (with Vamstar Ltd.) that developed text mining methods and solutions to mine large scale, heterogeneous, multilingual procurement documents in the healthcare sector (pharmaceuticals in particular), with an aim to construct a structured database of supplier contract histories. The database incorporates fine-grained supplier information that allows presenting a 'supplier risk profile' in terms of their capacity and credibility in fulfilling contractual terms. It also enables gauging regional supply chain capacities by aggregating individual supplier data. To the best of our knowledge, our work makes contributions to industrial text mining in several ways: 1) develops the first structured database for healthcare procurement that can facilitate the tendering process; 2) being the first to document the holistic backend text mining and NLP process involving multiple tasks working on heterogeneous, multilingual datasets. This is based on a method that effectively uses domain knowledge and can be easily generalised to multiple tasks and languages; 3) discusses lessons learned from adapting text mining research to developing real world applications in industrial contexts.

The remainder of this paper is structured as follows. We first (Section 2) introduce the domain and task studied in this work, setting the wider background for literature review, which is covered in the following section (Section 3). We then introduce our proposed solution (Section 4). Next in Section 5 we report evaluation of the components and present the end product. We further discuss lessons learned (Section 6) from this work and conclude with a reflection on the limitations and future work (Section 7).

## 2. Domain and Task

This work focuses on healthcare procurement, which has been rarely studied in the literature. The **primary goa**l **of the project** is to develop a platform that allows the dynamic creation of a 'supplier risk profile' for each healthcare supplier. We envisage such a profile to consist of different 'indices' that evaluate different perspectives (e.g, capacity for supplying certain products, geographical coverage) of 'risks' for potential buyers to sign contracts with the supplier. This would enable questions such as 'who are the suppliers able to supply this kind of medication', 'to what extent are they capable of supplying for this country', or 'are they able to supply such quantity' to be easily answered. Such questions are often crucial for buyer decision making. However, the current procurement process relies on manually sifting through multiple lengthy documents to seek answers. This is a very resource consuming process. Understandably, an enabler of our primary goal would be a structured database of healthcare suppliers' historical contract data. Thus **the secondary goal of the project** is to develop such a database and populate it with historical healthcare procurement data. While public procurement data is vastly available, as we shall explain in the following, there is a mixture of structured, semi-structured, and unstructured multilingual data that need to be mined and linked. Therefore, a major part of the project's work is developing text mining and NLP solutions that automatically process large quantities of unstructured procurement data to mine information that can be used to populate the database. **The goal of this article** is therefore, to report the development of these text mining and NLP methods.

### *2.1. Data sources and complexity*

The project targets procurement data from the 'Tenders Electronic Daily' (TED) platform, which is used by the EU governments to publish their public procurement related projects. TED publishes over 460,000 calls for tenders and contract awards in 26 official European languages per year, for about 420 billion euro of value. Each tender may be divided into multiple 'lots', where a lot is the smallest contract unit. Each lot may contain multiple items that are required. As an example, tender notice '2019/S 180-437985'[1] lists 47 lots from an NHS (UK) tender, with their sizes ranging from 2 to over 30 items. If a tender secures successful bids, a 'contract award' (or multiple awards) will be made and recorded in TED for the tender. In the following, for the sake of explainability, we assume there is one award for each tender (however in practice, our methods are applied to all awards that are available for a tender). Note the lots offered in a tender and the contract awards form a 'many-to-many' relationship. Namely, multiple lots can be awarded to a single entity and documented in one contract award; a single lot can also be awarded to multiple entities, forming multiple contract awards; further a single contract award can include one or multiple lots.

On TED, each tender and its corresponding contract award(s) has a structured XML file documenting key elements of information. We refer to these as 'tender XML' and 'award XML'. An example of a tender XML is shown in Figure 1. Award XMLs generally follow the same structure. Tender XMLs document information such as the buyer, the lots, items of lots, contract criteria, etc. Award XMLs document the buyer, the lots, the awarded suppliers for each lot, contract value, quantity, etc. Each tender may also have a collection of 'attachment documents' that provide further details of the tender, especially on lots and items ('tender attachments').

---

[1] https://ted.europa.eu/udl?uri=TED:NOTICE:437985-2019:TEXT:EN:HTML, last accessed: Nov 2022

```
11/03/2020    S50
I.  II.  III.  IV.  VI.
                        North Macedonia-Strumica: Pharmaceutical products
                                        2020/S 050-119757
                                         Contract notice
                                            Supplies

Legal Basis:
Directive 2014/24/EU

Section I: Contracting authority
    I.1)  Name and addresses
          Official name: Public Health Institute Hospital Strumica
          Postal address: Mladinska br.2
          Town: Strumica
          NUTS code: 00 Not specified
          Postal code: 2400
          Country: North Macedonia
          Contact person: PHI General Hospital Strumica
  ...   ...
    II.2)  Description
    II.2.1) Title:
          Gentamicin solution for inj. 40 mg/2ml
          Lot No: 94
    II.2.2) Additional CPV code(s)
  ...   ...
    II.2.5) Award criteria
          Criteria below
          Price
    II.2.6) Estimated value
```

Figure 1. Excerpt of an example tender XML from TED (notice ID 2020/S 050-119757). Note section II.2.1 lists a specific lot and its items, while II.2.5 lists the contract criteria.

Given the availability of tender and award XMLs, one may consider the task of developing and populating the database to be easy. However, the data in reality is far more complicated. **First** and foremost, the tender and award XMLs are often incomplete. The predominant missing information is lot and item information. As an example, the tender XML for '2019/S 180-437985', mentions 47 lots in the tender, without detailing the specific items but a lot reference number. This critical information is available from a bulk download of 7 tender attachments (PDFs). Both the tender and award XMLs then cross-reference these data sources through the use of the lot references. Recovering such information is crucial to building the supplier risk profile, which needs to account for the range and quantity of products that a supplier has supplied in the past. **Second**, not every tender attachment is relevant for our aim. Among those for '2019/S 180-437985', two PDFs list the actual lots and items (e.g., Figure 2), while others document specifications, requirements, regulations and protocols etc. **Third**, not every page of a relevant attachment contains relevant information. For example, Figure 3 shows that in another tender, lots and items are described in one page but different sections of a long document. **Fourth**, as it is already shown in Figures 2 and 3, there is a significant discrepancy in how lot and item information is described within the same country, or indeed, even the same organisation. This discrepancy has been observed at different levels such as: the use of structured formatting (e.g., free text v.s. tables/lists); the amount of information encoded (e.g., the table in Figure 2 lists 16 columns (attributes) for each item) even for the same kinds of products/services; and the semantics of structure where structures are adopted (e.g., the order and names of columns). Such a high level of complexity and inconsistency could be one major reason why there has been a lack of text mining and NLP studies or applications for healthcare procurement.

Figure 2. A snapshot of one PDF attachment that is part of the tender '2019/S 180-437985' (NHS, UK). The picture only shows some of the columns of the table, due to limited page space. Each row describes an item, while column 1 indicates lot references (as numbers).

Figure 3. An excerpt of one PDF attachment that is part of the tender '2020/S 111-270678' (Department of Health and Social Care, UK). The picture only shows part of a page of one PDF document that lists the lots and items. Pricing information is shown on other pages.

## *2.2. Task definition*

Relating to the project's ultimate goal of enabling the dynamic creation of 'supplier risk profiles', we envisage a database scheme that is partially (for reasons of both proprietary information and simplicity) depicted in Figure 4 showing entities (rounded boxes), their attributes (dashed boxes) and relationships (lines). Briefly, we hope to create a record for each contract award that has one buyer and one supplier, with associated lots (one or multiple) and item information. We store certain attributes of the buyer, supplier, contract award, lots and items, such that we can run queries to fetch 'supplier centric' award information like the range of products (based on lot items) they have supplied, the monetary value and quantities for different kinds of products, their buyers, and covered geographical regions (buyer country). Previously we mentioned that one award XML may mention multiple contracts with multiple suppliers. In practice, we 'flatten' data extracted from the award XML into multiple contract award records.

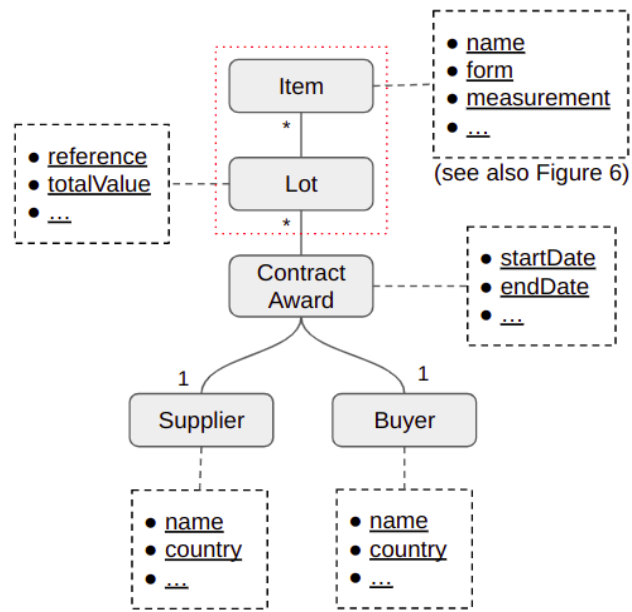

Figure 4. Partial database schema showing information of suppliers, buyers, contract awards, lot and item information to be extracted and stored. Round boxes indicate entities, dashed boxes list attributes in bullet points (only showing those important for understanding the rest of this article). The '1s' indicate there is one supplier and buyer in one award record; the '*s' indicate there can be multiple lots in an award record and multiple items in a lot.

As already mentioned, much of the information captured by this schema is already in (semi-)structured format available from the tender and/or award XMLs. These can be easily extracted through XML parsing, although some light-weight data cleansing and linking may be needed (e.g., extracting the country from a buyer/supplier address, or matching buyer names from the tender and award XMLs). These tasks are relatively straightforward and will not be the focus of this article. Instead, the real challenge is extracting and populating the lot and item information (indicated in red dotted box) that is often missing in both the tender and award XMLs. Therefore, our tasks can be defined as it is shown in Figure 5 - given a tender XML, its associated award XML, and the associated tender attachments, we aim to:

1. Extract the structured lot and item information often missing in tender and award XMLs (filling the schema highlighted in the red dotted box in Figure 4; also the middle lane in Figure 5):
    a. From the tender attachments, identify the relevant documents, and the relevant content areas (i.e., tables, lists, paragraphs) that contain information about the lots (lot zoning);
    b. From the identified content areas, detect content elements (e.g., sentences) associated with lot items (lot item detection);
    c. From the detected lot items, parse the lot and item descriptions into a structured representation, identifying attributes such as the lot reference, name of the items, measurement units, etc (lot parsing). As an example, Figure 6 shows how we want to extract structured lot and item information from the example in Figure 3.
2. Extract the structured supplier, buyer, and contract award data from the tender and award XMLs (filling other parts of the schema shown in Figure 4; also the upper and bottom lanes in Figure 5).
3. Join the information extracted from 1 and 2 above to form supplier-centric contract records and populate the database.
4. Develop supplier risk indices using the populated database and a platform for exploring these indices.

For the scope of this article, we will focus on 1), briefly cover 2) and 3) as they are more straightforward and completed outside the scope of this project, and skip 4).

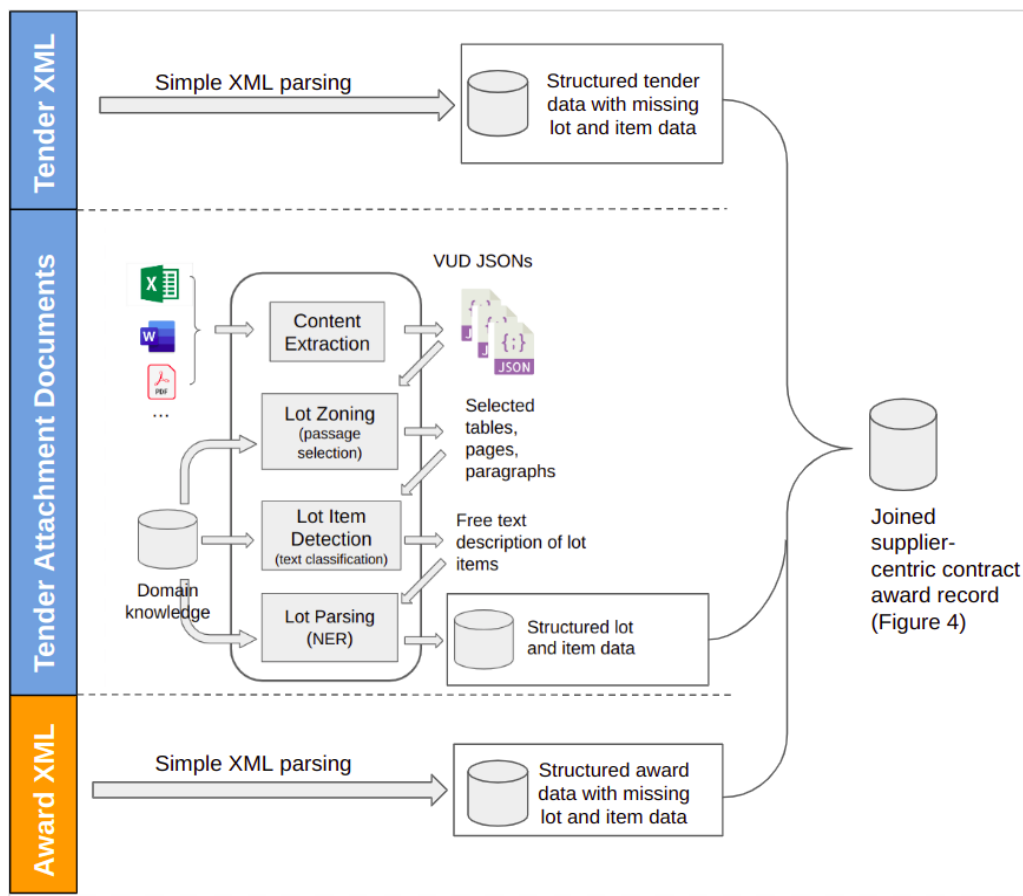

Figure 5. The overall workflow of the award record database construction process.

Figure 6. An example showing how structured lot information should be extracted.

## 3. Related Work

Due to the multi-task and heterogeneous dataset nature of our work, we can identify many areas of research that are relevant. Our literature review will focus on the following scope. First, we give an overview of the research of different text mining and NLP tasks relevant to our work. We avoid presenting details of individual studies due to the too wide scope and the massive amount of literature. Our intention instead is to draw literature that helped inform developing our solution and reflect on why mainstream research findings cannot be easily adopted in industrial projects. Second, we examine the areas where text mining and NLP are used to develop real world applications, such as the legal and construction domains mentioned above. Again we will focus on a 'big picture' with an aim to articulate the difference from the procurement domain. Finally, we give a detailed discussion of work done in procurement text mining, the core area that our work belongs to. Here we will cover details of each study and compare it against this work.

## 3.1. Text mining and NLP research overview

Text mining and NLP are two different, but overlapping research fields each encompassing a wide range of tasks. The first focuses on discovering or extracting information from unstructured texts, NLP on the other hand, focuses on enabling machines to understand human natural language. For example, named entity recognition (NER) and relation extraction (Han et al, 2020; Nasar et al., 2022) serve both purposes and therefore, can be viewed as both text mining and NLP methods. However, speech recognition is a subfield of NLP but not related to text mining. Both text mining and NLP are long-established research fields that witnessed decades of development and the creation of sub-communities looking at specific tasks. Due to the space limit of this article, we only discuss a number of tasks that are relevant to this work, including: text classification, NER, text passage retrieval, and table information extraction.

**Text classification** (Kowsari et al., 2019) is one of the earliest text mining and NLP problems and it has been widely studied and addressed in many real applications. In simple terms, the goal is to assign predefined categories to text passages. In their survey on text classification, Kowssari et al. (2019) identified four levels of text passages: full document, paragraph, sentence, or sub-sentence where segments of a sentence are examined. While the definitions of labels and text passages depend on domains, it is widely acknowledged that two key sub-processes are feature extraction and classifier training. Feature extraction converts texts into numerical vector representations for machine learning, and over the years has evolved from extracting lexical/syntactic patterns (e.g., bag of words, n-grams, words, part of speech) to word embeddings that are learned through modelling word contexts based on very large corpora (Birunda and Devi, 2021). When feature extraction results in high dimensional, sparse vectors, dimension reduction techniques (e.g., Principal Component Analysis, Latent Dirichlet Analysis) are used to transform a high dimensional vector into a low dimensional one optimised for machine learning. Following this, the classifier training stage attempts to learn patterns that can differentiate texts with different labels from their feature representations. This requires a dataset with ground truth labels to be provided (i.e., training data).

**NER** (Nasar et al., 2022) deals with the extraction and classification of mentions of named entities from texts. In 'He was named on the bench by *Mexico* in midweek but *Wolves* received an apology for that as *Raul Jimenez* was never going to play', texts in italics are NEs that are classified as country, organisation (football club) and person (footballer) respectively. NEs represent important information in a text and are often anchors for locating more complex information such as relationships and events. By definition, NER comprises two sub-processes that are often dealt with by supervised classification. The first is to identify 'boundaries' of an NER which can be a sequence of tokens. The second is classifying that sequence into predefined categories. Therefore, arguably, NER can be seen as low granularity text classification at token level and hence follows the same 'feature representation' and 'classifier training' steps. Features are often described at token-level, and can include for example, lexical/syntactic patterns of the token, and those of its preceding and following tokens, or word embeddings.

Both text classification and NER are relevant to our work as the first may help filter irrelevant documents or content areas through binary text classification, while the second allows targeting specific units of information (e.g., contract items, volume) that need to be extracted. However, we face unique challenges of no training data, and highly inconsistent content structure. Research in both fields are long-established, and predominantly report studies conducted with well-curated datasets in consistent format and structures. For example, multiple initiatives (Sang and Meulder, 2003; Piskorski et al., 2021) have been set up to foster NER research. But the training data are well-curated, free form sentences, even with raw input tokens and their features extracted and aligned. These hardly represent real-world problems. As discussed before, data we are dealing with are highly inconsistent in terms of their content formatting and structure, and we have no training data but general purpose domain lexicons. It is impractical to directly apply state-of-the-art methods in our scenario.

**Passage Retrieval** (PR) is often used as a subprocess in Information Retrieval and Question Answering (Zhu et al., 2021). The goal is to find matching textual fragments (or passages) for a given query, albeit at a lower granularity than document. PR deals with splitting a long document into passages, and scoring and ranking them against a query or question. In Othman and Faize (2016), common methods for each task are well-summarised. Briefly, passage splitting may be based on textual discourse units such as subsections and paragraphs, arbitrary windows that select a fixed or variable number of words/sentences, or 'semantic similarity' where units such as NEs are extracted from documents first and matched against the query, and then used as anchors to locate text passages. Ranking passages may be based on statistics such as the classic TF-IDF model used for document retrieval (Llopis et al., 2015), or machine learning that takes into account different features extracted from each passage.

PR is related to our work as part of our task is to identify text passages that contain relevant information for further analysis. However, our work is different in two ways. First, we do not have specific queries or questions to match against, but a rather vague notion of 'relevance'. Second, we need to identify multiple passages, while in typical PR for IR or QA, a single best match is needed.

**Table Information Extraction (TIE)** aims at extracting data points from tabular content and recovering the semantic relationships between data points. This includes a very wide range of tasks that can be generally divided into those detecting and parsing table structures to understand the geometric relationships between table elements, and those conducting semantic analysis of its content to interpret the relationships embedded in the textual content. A recent survey (Zhang and Balog, 2020) covers some of these tasks, while here we only explain the tasks relevant to our work briefly. **T**able **S**tructure **P**arsing (Paliwal et al., 2019) usually starts with images (including PDFs where content is not machine accessible) that contain tabular data, to detect tables, locate their cells, and analyse the row-column relationship (including complex column/row spans). **T**able **I**nterpretation deals with identifying entities from table text, and their semantic relations encoded by the table structure.

Both tasks are relevant to our work. As mentioned above, a significant portion of the relevant content in our procurement data is encoded in tables, which are predominantly found in PDF documents. But the structure of tables varies depending on countries and contracts. Therefore, TSP is essential for detecting such content, parsing the complex tabular geometric relationships, and converting the data into a structured, machine accessible format. However, tables can include irrelevant data, or they can encode certain types of content and relationships in different ways (e.g., different pairing of columns). Hence TI is needed to interpret the semantic meanings of table columns and rows, to allow extracting the only relevant information into structured formats. As we shall explain later, while we are able to use state-of-the-art TSP tools to detect and extract tables, we have to opt for a different approach to TI. This is because typically, TI depends on an external knowledge base that provides essential information for interpreting the content and their semantics in a table. In our case, such a KB does not exist.

### 3.2. Text mining and NLP in industry use

Text mining and NLP already have wide use in a number of industry contexts. Here, due to limited space, we only look at a few domains briefly and focus only on work for real applications instead of general purpose tasks such as building domain corpora or language models, or fundamental NLP research adapted to domain specific data.

In the **legal domain**, Zhong et al. (2020) summarised three main application areas: judicial decision prediction, similar case matching and question answering. Judicial decision prediction studies the problem of determining the verdict of a court on the basis of textual information about a court case before the verdict was made. This is typically treated as a text classification task, where court decisions are categorised and they are predicted based on features extracted from the case texts. For details, we refer readers to a survey by Francia et al. (2022). Similar case matching is an important application where judicial decisions are made according to similar and representative cases in the past. The goal is to find pairs of similar cases. This is often cast as a retrieval task, and is being extensively studied in Legal IR (Xiao et al., 2019). Similar to judicial decision prediction, existing methods mainly focus on comparing case texts at word or semantic level. In both areas, recent studies have argued for extracting more fine-grained 'case element' information (e.g., through using NER) that may better represent legal cases. For example, Hu et al. (2018) extracted 'legal attributes' that define the nature of cases in judicial decision prediction. Question answering is a rather complicated NLP task that builds on many low-level tasks, such as finding relevant text passages (PR and text classification), locating case elements (NER and relation extraction), and text similarity matching. QA is not related to our work. For both reasons, we do not go into detail about legal QA. But generally, questions typically concern explanation of legal concepts or case analysis.

Legal documents are often lengthy, but well-structured into different parts and follow standard structures that make them easier to process. Over the years, the research community in this domain has defined granular tasks, standards, and created rich resources including training data. In contrast, our work deals with much more heterogeneous data and inconsistent structures, where creating training data for some tasks is expensive. Most established methods in this domain are therefore not directly transferable to our task.

With the growth of e-commerce and social media, mining Web content related to **service/product provisions** has been extensively studied and applied to develop competitive advantage for businesses. A major area of application is analysing reviews from e-commerce or social media platforms to acquire business intelligence that inform various practices. For example, through analysing customer reviews about their own services or products, one can gain insights on customer satisfaction, to inform public relation management (Nave et al., 2018) and product development (Yang et al., 2019). Analysing such data about a sector in general including multiple competitors allows discovery of market trends and even forecast of demands (Chatterjee, 2019; Sharma et al,. 2020). Further, with the increase of platforms of user-generated content, the diffusion of fake information is made easier and becoming an increasing concern. Therefore, work has been done to automatically detect and filter such misinformation (Fang et al., 2020). A thorough review of work in the above areas can be found in Kuma et al. (2021).

Work in these areas heavily rely on sentiment analysis, which is a type of text classification task aiming to analyse the sentiment of a text. It is also very useful to extract key elements related to service/product provision (e.g., product features, service processes) for more fine-grained analysis, and this can benefit from NER methods. Many studies also use topic modelling and social network analysis, which are beyond the scope of this work. Compared to the domain in our work, data studied in these areas are rather homogeneous - they are typically free-form review texts that are independent and self-contained, making it easy to adapt state-of-the-art methods. Partially for this reason, there is an abundance of tools developed in this area. However, as explained before, the data we have to deal with is much more complex.

Text mining and NLP are also widely used in **system requirements analysis** and **quality control in construction**. To name a few, Li et al. (2015) processed software user manuals (PDFs) to extract functional and non-functional requirements. A dictionary of keywords combined with rules are used to extract sentences and topic modelling is applied to group similar sentences into groups, which may represent the same/similar requirements. The work also needed to deal with filtering irrelevant text passages in the extracted texts from PDFs, but this was done manually. Our work is similar in that some of our methods make use of keywords and rules, and also process PDFs. However, we need to deal with a much larger and more heterogeneous collection and the filtering of content must be done automatically. Tiun et al. (2020) developed supervised classifiers of functional and nonfunctional system requirements using well-curated training sentences obtained from Siemens Logistics and Automotive Organization. Khan et al. (2020) analysed Reddit posts about Google Maps, and trained a text classifier to discover posts related to the software feature requirements or issues using a set of manually annotated posts. Both deal with higher quality of data and have access to a large amount of high quality training data, while our project only managed to create limited training data for some but not all tasks due to resource constraints.

Lee et al. (2014) applied keywords extraction and co-occurrence analysis in marine structure quality inspection reports. The co-occurrence map is used to aid humans in identifying which quality aspects are most frequently mentioned and in what ways. Similar methods are used in Tian et al. (2021), who processed construction project reports to extract keywords and concepts related to several key aspects of project management (e.g., quality control, safety management). Zhang et al. (2019) developed a text classifier to classify accident reports from construction projects, and implemented a rule-based extractor to identify causes from accidents. The data used in these studies are also rather simple compared to ours. For example, report cases used by Zhang et al. (2019) are composed of short paragraphs and use a consistent structure.

### 3.3. Text mining and NLP for procurement

Compared to other domains, there is a lack of work on text mining/NLP for procurement.We discuss a few in this section. Grandia and Kruyen (2020) conducted an exploratory study of over 140,000 procurement notices from the Belgium E-procurement platform between 2011 and 2016, to analyse the trend towards 'sustainable public procurement' (SPP). The work does not extract any information from procurement documents. Instead, a light-weight keyword matching process is conducted to search for and count keywords related to the SPP concepts found in the procurement documents. To compile the keywords, they manually studied a large collection of legislations, policies and white papers. The keywords are then manually grouped into different themes (e.g., circular economy, social return), and matched to the text content inside procurement documents for counting. Authors also mentioned the heterogeneous nature of procurement documents - although each record has a compulsory XML document, very little useful information is encoded in it but inside a vast collection of PDF, Excel, or Word documents that need to be parsed to a machine readable format. Haddadi et al. (2021) later applied a similar process to the ICT sector in Morocco to measure, also to gauge the trend of addressing 'sustainability' in public procurement. Both studies do not extract structured information from procurement documents but opted for simple keyword matching. This does not require complex preprocessing or interpretation of document structures in order to target the right content areas. Our task in comparison, is much more challenging.

Chalkidis et al. (2017) is one of the earlier studies that looked at extracting structured information from contract data. They developed an NER process that identifies different elements (title, parties, start and termination dates, contract period and value) in a contract document that is already converted to free-form texts. Using a well-curated dataset of just over 2,400 contracts (sources unclear), the model is trained using features such as word casing, token type (number, letter), length, and part of speech etc. Authors also used post-processing rules to fix boundary detection errors (e.g., '2013' missing from the date '23 Oct 2013'). Our work also uses a mixture of supervised and rule-based methods. But the fundamental differences are that we have to deal with multiple tasks beyond just NER, we do not have well-curated training data for all tasks, and our datasets are much more complex.

Choi et al. (2021) analysed engineering and construction contracts with a goal to extract risk phrases and entities using rule-based phrase matching. Authors acknowledged the content accessibility issue with procurement documents, which are typically PDF documents. They used an OCR process that recognises sentence boundaries, and opted for a sentence classification solution. Each 'risk factor' is associated with a domain lexicon, which is used to match and label sentences. Our work is similar in the way that we also make use of domain lexicons. However, we use both rule-based and self-supervised methods, and we generalise the method to a number of NLP tasks and different languages.

Fantoni et al. (2021) working on the railway domain, acknowledged the heterogeneity and complexity of procurement documents - some large engineering systems may list over 100,000 requirements, documented in inconsistent ways and multiple documents that use non-standard structures and layouts. Their goal is to 1) identify from multiple, long procurement documents the sentences describing system requirements; and 2) classify these requirements into different railway subsystems/components. Authors mainly used unsupervised methods. Starting with building a domain 'knowledge base', they used keyword/phrase extraction methods to build a lexicon specific to the domain. Then rules are developed to match low granular information equivalent to named entities, such as measurement units, standard references etc. This extracted information is later used to match sentences in the document, and a score is computed based on the number of matched units and the subsystems/components they relate to for ranking purposes. Again our work also uses domain lexicons but we apply them to both rule-based and supervised methods for multiple tasks.

Rabuzin and Modrusan (2019) highlighted a lack of studies of text mining and NLP in procurement analysis. With a goal to extract technical conditions and criteria in procurement contracts, they downloaded documents from the Croatian procurement portal and converted them into machine readable texts. Authors acknowledged the complexity and inconsistencies in document structures, hence the challenge in identifying the relevant content areas for further analysis. They opted for a simple 'sliding window' solution: a 1000-word window around occurrences of 'technical' and 'professional'. The task is then transformed into a text classification one, for which authors trained three algorithms for comparison. However, it is unclear how the training data is created. The authors later updated their work (Modrusan et al., 2020) in terms of how the sliding windows are defined. Compared to our work, these studies belong to the task of 'passage retrieval', while we deal with not only PR, but also multiple, more fine-grained extraction tasks.

### *3.4. Conclusion from literature review*

Our literature review shows that, despite significant research in the areas of text mining and NLP, there is a strong dominance by supervised methods built on well-curated data that do not transfer well to practical scenarios. This is partially reflected by the number of industrial text mining/NLP studies that incorporated rule-based methods and the use of domain lexicons, except a few areas (e.g., the legal domain) where high quality curated resources are abundant. The majority of industrial studies also look at single and sometimes simplified tasks, but do not report a full process in an end-to-end fashion, particularly with a lack of details on how data heterogeneity and inconsistency is dealt with by their methods. Further, no prior work has focused on the healthcare domain. Our work will address these gaps.

## 4. Proposed Methodology

Figure 5 presents an overview of our workflow. As mentioned before, this article focuses on extracting the structured lot and item information often missing in tender and award XMLs (the middle lane). This will be covered in Sections 4.1 to 4.5. In Section 4.6, we briefly cover the other parts of the workflow.

Given a collection of tender attachment documents associated with one tender notice, our first step (content extraction) is to use various data extraction libraries to convert various content formats into a single, universal data structure called 'Vamstar Universal Documents (VUD)' that represent text content in JSON format. In 'lot zoning', we use passage retrieval/selection techniques to identify the content areas (pages and tables) that potentially contain useful lot information. Next, 'lot item detection' uses text classification techniques to process the extracted text passages to identify content (sentences and table rows) that describe a lot and its items. Following this, we apply rule-based NER to parse the texts related to a lot and its individual items to identify specific attributes and create a structured representation of the lot (lot parsing). For most of these processes, we use domain knowledge in a generalisable way for multiple languages.

Meanwhile, other structured information is extracted from tender and award XMLs by simple XML parsing. These extracted information is then joined to form supplier-centric contract award records, which are used to populate our database. The database is then used to calculate supplier risk indices, which form a supplier risk profile. In the following sections, we explain each component. However, details of certain components may need to be redacted due to NDAs on proprietary content.

### 4.1. Domain knowledge

With the semi-structured data available from TED tender XMLs and award XMLs, Vamstar developed XML parsers that extract different fields and store their unique values in a temporary data store (to be called the 'tender and award fields' or just 'fields' in short). As an example, given the XML in Figure 1, a record would be created with fields such as: buyer name and addresses, lots (e.g., 'Verapamil solution for inj. Lot No: 62' but with multiple values as the notice contains more than 90 lots), award criteria, etc. This allows generation of useful domain knowledge in different ways.

First, **training data** for different fields can be created by simply taking the unique values from each field. These values usually take the form of short phrases or sentences, and therefore, can be used to train text classifiers at phase/sentence level. Note that this data is multilingual.

Second, **domain lexicons** for different fields can be built by extracting 'representative' words from each field. Given a field that we are interested in, we merge all values from different languages, and use machine translation tools to translate them into English. Instead of drawing a boundary between what is and is not a 'representative' word for that field, here we explain a process to calculate the 'specificities' of words and we refer to this as 'domain specificity lexicon', which is used in other components of our method. Given a word $w$, the basic idea is to measure how unique $w$ is to lot/item descriptions compared to other text content in a tender. The basic process is as follows:

1. Identify the set of 'fields' that are most relevant to a tender. For this we selected 5 fields - this is rather a subjective interpretation by the domain experts: 'Name and addresses' of buyers (Section I.1.1 of the notice in Figure 1), 'notice title' (II.1.1), 'short description' (II.1.4), 'lot and item descriptions' (a concatenation of values from fields like II.2.1), 'contract criteria' (a concatenation of values from fields like II.2.5). Following the process mentioned above, all values are then translated into English;
2. For each field, create a bag-of-words ('BOW') representation by concatenating the values of that field across all records in the database, applying stop words removal and lowercasing;
3. For $w$, calculate 4 metrics as follows: $ntf(w)$ normalised frequency of $w$ in the BOW representation of the 'lot and item descriptions' field, calculated as the ratio between the frequency of $w$ in the BOW over the total words in the BOW; $ndf(w)$ the normalised 'document frequency' of $w$, calculated as the ratio between the number of fields in which $w$ is found over the number of all fields (i.e., 5); $\overline{ntf \cdot ndf(w)}$, inspired by the tf-idf measure used in document retrieval, this is the product of $ntf(w)$ and the inverse of $ntf(w)$; $weirdness(w)$, a score calculated by comparing $ntf(w)$ against that calculated in a reference, general purpose corpus (in this case, the Brown corpus).

Thus our domain specificity lexicon contains unique words found within the 'lot and item descriptions' field, and each word has four associated scores. In addition to this, we also use another two dictionaries compiled by Vamstar. One is a list of words often used as measurement units (e.g., mg, ml), and the other is a list of words often used to describe the 'form' of required items (e.g., pack, box, bottle). Both are very short and contain only a few dozens of entries. These lexicons and dictionaries are also translated to other languages using Google Translate.

### 4.2. Content extraction

In this component, our goal is to convert heterogeneous data file formats (Word, Excel, PDF, etc) into a universal, machine accessible JSON format, which we refer to as VUD. For each file, depending on its format, we use the corresponding APIs (e.g., Apache Tika for Word and Excel files, Apache Tesseract and PDFPlumber for PDF). However, not all APIs or data files support the extraction of formatting features (e.g., font size, colour, header level), especially if a document is not structurally tagged. Therefore, we focus on extracting only the text contents and a limited range of structure information. Our VUD documents contains the following structured content:
- Pages, corresponding to the pages in the source document;
- Paragraphs, identified as a consecutive block of texts separated by at least two new line characters. E.g., in Figure 1, 'North Macedonia-Strumica: Pharmaceutical products' and '2020/S 050-119757' in the title will be treated as separate paragraphs, while 'Legal Basis: Directive 2014/24/EU' is a single paragraph;
- Tables, extracted using the Camelot and Tabula libraries, contain basic tabular structures including column and row indices (including column and row span information), and the textual content within each cell. Tables across multiple pages are recognised as separate tables initially, but are then merged if they meet the following simple rules: 1) there are no other non-tabular structures between two tables; 2) they have the same number of columns.

### *4.3. Lot zoning*

The text content extracted above may or may not contain relevant information for lots. Therefore, this component deals with the filtering of the text content to identify the relevant content areas for further processing (i.e,. 'zoning' to the right content). We focus on two tasks: 1) selecting the relevant pages; 2) selecting the relevant tables. We define 'relevance' as containing a description of a lot (lot reference) and/or its items (lot items). And for both, we cast them as a text classification task. For example, in Figure 2, a valid 'lot reference' is 'Lot Number 1', and it contains two items (row 2 and 3); in Figure 6, the first bullet point in Section 2.1 has a lot reference 'Lot 1', with the following text being its only item.

### *4.3.1. Page selection*

Given a page $p$, we define $s \in p$ as a sentence from $p$, and $BOW$ as a function that can be applied to $s$ or $p$ to convert it into a simple BOW representation applying stopwords removal and lowercasing (lightweight NLP that is language dependent). We create several language-independent features for learning a relevant page classifier such that it can be used for content in any language.

For each word $w$ in the domain specificity lexicon $w \in L$, we search the word in $BOW(p)$ and count its frequency as $count(w, p)$. Then for each of the four metrics introduced above, we create four features (so a total of 16 features): the sum, average, minimum and maximum of the scores, as follows, where '*metric*' generalises the four metrics explained before (*ntf, ndf, ntf-ndf, weirdness*). Further, for each of the two domain dictionaries (form and measure), we create four features (so a total of 8 features) using the same equations below, by replacing $L$ with a dictionary $D$, and setting $metric(w) = 1$ always. Together this gives us a total of 24 'page-level' features.

$$feature_{sum}(p) = \sum_{w \in BOW(p) \cap L} count(w, p) \times metric(w) \quad [1]$$

$$feature_{avg}(p) = \frac{feature_{sum}(p)}{\sum_{w \in BOW(p) \cap L} count(w, p)} \quad [2]$$

$$feature_{min}(p) = min_{w \in BOW(p) \cap L} \{count(w, p) \times metric(w)\} \quad [3]$$

$$feature_{max}(p) = max_{w \in BOW(p) \cap L} \{count(w, p) \times metric(w)\} \quad [4]$$

While the above features are based on word frequencies within a page without considering sentence/paragraph boundaries, we also calculate 24 sentence level features. Using equation 1 as an example, if we replace $p$ with $s$, we calculate the same metric for a sentence $feature_{sum}(s)$. Then we define another feature $feature_{sum}(s, p)$ as the sum of $feature_{sum}(s)$ for all sentences in $p$. Similarly, we calculate $feature_{avg}(s, p)$, $feature_{min}(s, p)$ and $feature_{max}(s, p)$. These measure the density of the domain specificity lexicon used at individual sentence level. Replacing *metric* with the four options, we obtain 16 sentence level features. In the same vein, replacing $L$ with the two dictionaries $D$ and setting $metric(w) = 1$ gives us another 8 sentence level features.

Finally, we calculate the percentage of sentences that contain at least a number (e.g., indicating a volume/quantity or measurement although can be noise sometimes), and evaluate the percentage of sentences that contain at least one matched word from the lexicon/dictionaries and are calculated as follows (where $L$ indicates the domain specificity lexicon but can be replaced by either of the two dictionaries):

$$feature_{sent}(p) = \frac{\sum_{s \in p} 1\,[BOW(s) \cap BOW(L) > 0]}{\sum_{s \in p} 1} \quad [5]$$

In total, we obtain 52 features for a page. Our feature extraction process is arguably language independent, as it depends on counting word frequencies using lexicons/dictionaries and pre-computed scores assigned to the entries in these lexicons/dictionaries. The language dependent processes are the translation of the domain lexicons/dictionaries, and the lightweight NLP processes for which models for different languages are more available. For businesses, these help create more affordable multilingual NLP solutions than other choices such as translating input documents, or using more language dependent models. We will discuss this further in later sections.

In order to learn a classifier that determines if a page is relevant or irrelevant, we asked domain experts to annotate a random collection of pages extracted from the raw TED datasets. For the machine learning algorithms, we compare linear SVM, logistic regression, and Random Forest.

*4.3.2. Table selection*
Similar to page selection, the goal here is to filter tables that are irrelevant. This is common because tables are used in tender notices to describe not only the lots, but oftentimes award criteria, product specifications, and formatting purposes. In this work, we assume that all tables are 'horizontal', that is, their columns define attributes of data instances and their rows contain individual data instances. We define $t$ as a table and $r \in t$ and $c \in t$ represents rows and columns in the table. If we consider content in a table as a structure-less page, and concatenate each of its rows or columns into a sequence of words equivalent to a sentence potentially describing an instance, then we can apply the same feature extraction methods explained above. In cases of a row or column span, we duplicate the value across all the rows/columns covered by the span. Therefore, we obtain a total of 80 features as follows:
- 24 'page' level features if we treat the table as a simple flat textual page
- 28 'sentence' level features if we treat the table as a simple flat textual page, and its rows as sentences
- 28 'sentence' level features if we treat the table as a simple flat textual page, and its columns as sentences

For training, our domain experts annotated a collection of tables extracted from the raw TED dataset (to be detailed later) and we compared the same group of machine learning algorithms.

## *4.4. Lot item detection*
With the relevant pages and tables identified, our next step in the process is to identify the text elements in the pages/tables that actually describe lots and items. We would like to capture two kinds of information: the texts that indicate a lot reference and the texts that describe individual items in a lot. In practice, this is often contained in a single, coherent text passage, such as that shown in Figure 5. Thus we handle this as a binary sentence classification task where a sentence is a positive example if it contains either a lot reference, or item information, or both; and negative otherwise. We deter the extraction of lot reference and structured item information to the next component 'lot parsing'. We also need to deal with unstructured texts from a page and content from tables in slightly different ways.

*4.4.1. Unstructured texts in a page*
For unstructured texts in a page, we apply sentence splitting and then sentence classification. For features, we adapt the 24 page-level features explained above in the 'page selection' section by applying them to each sentence to classify. As an example, in equation 1, we replace $p$ with the input sentence $s$ and by alternating the $metric$ we obtain 4 features. The same can be applied to equations 2~3. Then replacing $L$ with the other two domain dictionaries will produce another 8 features. Further, we add three binary features that help capture lot references:
- If the sentence contains the word 'lot' (and its translation in other languages)
- If the sentence contains the word 'number' or 'no.' or 'num' (and its translations)

- If the sentence contains number-like tokens (including roman/arabic numerals and patterns such as '1.1, 1.21, X.2, II.1')

We also compare the same set of algorithms for classification.

*4.4.2. Tables*

Similar to the idea explained in passage selection, we simply treat each row as a sentence and therefore, the solution would be almost the same as that for dealing with unstructured texts above. One exception is for tables like that in Figure 2, where the lot references need to be interpreted based on combining the table header 'Lot Number' and the text in the row spans below that header. Thus for tables, we add additional rules for converting rows into sentences:

- We apply a rule to match the table headers against the pattern 'lot [token]' where [token] is an optional word (e.g., 'number', 'no.');
- For the column headers matched by this rule, we find the one where all the texts of cells in that column are 'number-like' tokens. Then, we insert the header text before the number-like token within that cell. As an example, the row span with a value '1' in Figure 2 will be updated to 'Lot Number 1' and will be repeated for both row 2 and 3 when creating a sentence.

Some may argue that lot items detection from tables should be simpler as once a table is classified as a 'relevant' one containing lot and item descriptions, given our 'horizontal table' assumption, we can simply take each row (after 'expanding' row/column spans) as an instance of item description. While this may be true for many cases like that shown in Figure 2, practically, there are complex table structures such as that shown in Figure 7. Here, each lot is split into a consecutive number of rows, and within each lot, the headers are repeated for the item(s) in that lot. The actual content we wish to extract from the tables are those indicated in the box. For this reason, we opt for the generic approach described above, i.e., treating each row in a table as a sentence for classification. The same feature set and algorithm described above are used, and the training datasets will be explained later.

Figure 7. A complex table containing lot and item information written in Portuguese.

## 4.5. Lot parsing

The goal of this component is to take the sentences (including those converted from table rows) classified as containing lot references/items (positive sentences), and parse them to create a structured representation of the lots/items such as that shown in Figure 6. There are two tasks in this process: 1) determining the boundaries of lots and the lot references, as the previous steps only classifies a sentence/table row as if it contains lot information; 2) from a positive sentence, extracts the structured item information including: name of the item, form, and measurement.

For 1), we use simple rules as follows:
- Apply the pattern to match the start of a sentence 'lot [token]+ [num]', where [token] is an optional word (e.g., 'number', 'no.') and [num] is a number-like token. If a positive sentence matches this pattern, it is considered to contain a lot reference and the [num] value will be extracted as the reference;
- Each sentence the above rule marks the boundary of a lot;
- Positive sentences including/following an identified lot reference are considered to be associated with that lot and are subject to the second part of processing.

As examples, in Figure 2, we expect rows 2 and 3 to be considered as part of a single lot with a reference indicated in column 1, and each row to describe a separate item. In Figure 6, we expect each bullet point in Section 2.1 to be classified as a relevant lot item, with the lot reference extracted as 'Lot 1' for the first and the remaining text for further parsing. In Figure 7, we expect the rows in the first and second boxes to be classified as relevant, while the first row identifies the reference and the second describes the item.

For 2), in theory, one can build an NER tool given sufficient training data. However, creating training data at token level is very time-consuming and deemed infeasible by Vamstar. Instead, Vamstar built an in-house rule-based tagger that uses a combination of regular expressions and dictionary lookup. Details of this part of the system is proprietary information and cannot be included in this article. However, we explain the main principles below.

First, using all values extracted from the 'lot and item descriptions' field of the TED database mentioned before, n-grams are extracted and their frequencies are generated. Domain experts were then involved in manually inspecting the derived ranked list of n-grams, to determine a 'threshold' (practically, different thresholds were derived for different 'n's) above which n-grams were deemed to be more reliable and retained as a lookup dictionary - to be called the 'lot and item n-gram dictionary'. Second, this dictionary and together with the 'form' and the 'measurement unit' dictionaries mentioned before are used to match texts within a sentence. Finally, once the matching is completed, post-processing rules will be applied to address overlapping boundaries, or extract other values such as quantities. For example, given 'Atropine Sulfate Solution for Injection 3 mg/10 ml Pre-filled Syringe', the 'matching' phase may identify 'atropine sulfate' and 'sulfate solution' as candidate n-grams, 'mg' and 'ml' as measurement, and 'syringe' as form. The post-processing will merge 'atropine sulfate' and 'sulfate solution', and process a context window of the identified measurement units to extract quantities.

### *4.6. XML parsing, data joining, and risk indices development*

So far, we have explained how we extract missing lot and item information from tender attachment documents. As per Figure 6, our workflow also parses tender and award XMLs to extract other structure information about contract awards. This information is then joined with the lot and item information to create supplier-centric contract award records. This is a relatively straightforward process that we will only briefly explain here. For each award XML, one contract award record is created for a unique supplier-buyer pair. The supplier information is extracted from the award XML; while the buyer in the award XML is matched to that in the tender XML through name matching. This allows us to integrate additional buyer information from the tender XML. Contractual terms such as the start and end dates and lot quantity and values are also available from the award XML. However, where the award XML has missing lot and item information, we use the lot references to map to the extracted lot and item information from the tender attachment documents. Through this process, we populate our database with the schema partially shown in Figure 4.

Using this database, we can retrieve a particular supplier and its contract history for particular products. We can also use this information to calculate quantitative 'indices' that summarise individual suppliers' risks from a buyer point of view. We introduce a number of metrics we developed in the current proof-of-concept. Once again, these are proprietary and therefore we focus on explaining the intuitions behind instead of revealing the mathematical calculations.

In total, we define 21 metrics that can be organised from two angles. For the first, we divide metrics into two types: one that measures a supplier's ability to supply different ranges of products, the other that measures a supplier's economic risk, taking into account their historical contract capacity and currently 'active' contracts. For the second, we define five sub-groups: 'Product Metric (PM)', 'Buyer Metric (BM)', 'Location (LocM)', 'Lot Metric (LM)', and 'Value Metric (VM)'. PMs measure the ability of a supplier to supply a number of different products, or indications of 'divergence' of their portfolio. These monitor each year's performance by taking into account their product portfolios and contract fulfilment track record. Factors included are, for example, the number of different types of products they supply each year; how this changes over time (increasing, or decreasing to a smaller number of products); quantity of specific types of products delivered to buyers.  BMs measure the extent to which suppliers are generally able to retain their customers (e.g., churn and retention). LocMs measure the supplier's ability to sell goods across countries and therefore, considers geographical coverage by their historical contracts. LMs measure at tender lot level the suppliers' ability to participate in multiple contracts over time. This accounts for contract durations and helps us infer supplier stock levels. VM measures financial capacity and different value propositions suppliers can offer and participate in. For example, these may look at the fluctuations in a supplier's contract capacity to identify risks of inability to supply. When a supplier signs a contract with a value (or quantity) that is significantly beyond its 'norm' (e.g., based on the average and range of contract values). The intuition is that signing contracts significantly deviating from the 'norm' can represent a risk of inability to supply. We will demonstrate the end system that allows exploring these risk indices and supplier profiles in the next Section.

## 5. Experiment and Demonstration

In this section, we present evaluation of some of the components above and demonstrate part of the end system.

### 5.1. Component evaluation

Here, we present evaluations of lot zoning and lot item detection, for which our domain experts were able to create some gold standard for model building and evaluation. In both tasks, we use text classification and compare three machine learning algorithms: linear SVM, logistic regression (LR) and random forest (RF). Therefore, we use the standard evaluation metrics for classification: Precision, Recall and F1. These are calculated as follows. Since each task has essentially two classes, we measure the scores for each class and take their average for reporting (macro-average).

$$Precision = \frac{\#True\ Positives}{\#True\ Positives + \#False\ Positives} \quad [5]$$

$$Recall = \frac{\#True\ Positives}{\#True\ Positives + \#False\ Negatives} \quad [6]$$

$$F1 = \frac{2 \times Precision \times Recall}{Precision + Recall} \quad [7]$$

To evaluate lot zoning at page and table levels, we created two separate datasets. We asked domain experts from Vamstar to annotate a random collection of pages and tables extracted from our raw TED datasets into 'relevant' or 'irrelevant'. This gave us a total of 781 pages, and 515 tables (multilingual in both cases). To evaluate lot item detection, we conducted two sets of experiments. First, using the TED database created by Vamstar, we identify the field containing values most similar to the description of lots and items, i.e., 'lot and item descriptions', split the values based on sentence boundaries and take the unique values. Values with less than 2 words or more than 20 words are excluded. This gives us a total of 18363 positive examples (multilingual). For negative examples, we apply the same process to the other four fields: 'name and addresses' of buyers, 'notice title', 'short description', and 'contract criteria'. This gives us a total of 15632 negative examples. Combined together, we refer to this dataset as 'lot item gold standard'. However, this only allows evaluating sentence-level classifiers, and the dataset does not necessarily represent the real documents that may describe lots in free-text and tables. Therefore, our second evaluation involves domain experts manually verifying the extractions (sentences, table rows) of the best model from 20 English tender notices for precision only. We refer to this dataset as 'lot item verification'. Statistics of these datasets are shown in Table 1. Tables 2 ~ 3 show the evaluation results for different tasks respectively.

| Task/Dataset | Positive examples | Negative examples |
|---|---|---|
| Lot zoning - pages | 479 | 302 |
| Lot zoning - tables | 158 | 357 |
| Lot item GS | 18363 | 15632 |
| Lot item verification | 897 | N/A |

Table 1. Dataset information for different task evaluation.

| Lot zoning - Pages | | | | Lot zoning - Tables | | | |
|---|---|---|---|---|---|---|---|
| Models | Precision | Recall | F1 | Models | Precision | Recall | F1 |
| SVM | **0.90** | 0.90 | 0.90 | SVM | 0.86 | 0.76 | 0.79 |
| LR | **0.90** | 0.90 | 0.90 | LR | **0.90** | 0.86 | **0.88** |
| RF | **0.90** | **0.91** | **0.91** | RF | 0.86 | **0.88** | 0.87 |

Table 2. Evaluation results for lot zoning. Best results for each metric are highlighted in **bold**.

| Lot item GS | | | | Lot item verification - best model only | |
|---|---|---|---|---|---|
| Models | Precision | Recall | F1 | Model | Precision |
| SVM | 0.89 | 0.88 | 0.88 | RF | 0.84 |
| LR | 0.90 | 0.89 | 0.90 | | |
| RF | **0.91** | **0.94** | **0.93** | | |

Table 3. Evaluation results for lot item detection. Best results for each metric on the lot item GS are highlighted in **bold**.

The results show that arguably, random forest is the most robust algorithm as it performed best on page-level lot zoning and lot item detection, while achieving comparable results on table-level lot zoning. We also conducted error analysis to understand where the automated methods failed.

In terms of the two 'zoning' tasks, we found three main causes that applied to both pages and tables. First, the low recall is typically due to 'sparse' signals, such as a tender containing only one lot with very few items, described in a very short table or paragraph/list. Content as such is difficult to extract because the features generated for those text passages would be sparse compared to 'denser' pages or larger tables. Low precisions are primarily due to 'noise' in the domain lexicon, as some generic words (e.g. 'solution', 'wrap', 'free' as in 'sugar free') were retained and scored high. To address these issues, domain experts suggested developing 'post processing' rules to 'recover' passages that are potentially valid candidates for processing. For example, enforcing that at least one text passage must be extracted from each tender (thus those not making the classifier decisions may be recovered). Also, excluding unigrams in the domain lexicon could help address false positives. The third common cause of errors is due to content extraction from PDFs. As an example, arbitrary whitespaces are recognised between every letter when parsing some PDFs (e.g., 'book' becomes 'b o o k'). This is an issue that is much harder to address, as it depends on the quality of PDFs as well as the third party extraction tool. This is an example of the practical challenges that industrial text mining and NLP projects can face while research may not as they deal with much higher quality data. This implies that in reality, system performance may be limited by some obstacles that are hard to overcome.

In terms of the lot item detection task, when examining the 'lot item verification' data, we find that the most common issue causing the decrease in precision is 'noisy' domain lexicon entries, as explained above. Further, complexity in the domain vocabulary also made the task challenging. For example, pharmaceutical ingredients sometimes have long names and are written in certain patterns, such as '17-alpha-hydroxyprogesterone'. For cases like these, domain experts suggested developing certain 'regular expressions' incorporating domain dictionaries. This implies that industrial text mining and NLP projects are often more complicated than tweaking a machine learning based method for performance measured by established standards. Practical solutions often rely on incorporating domain knowledge in various forms, oftentimes rules. This is consistent with earlier findings such as Chiticariu et al. (2013).

*5.2. System demonstration*

In this section, we present the end system to demonstrate the 'supplier risk profiles' in action. First, informed by the evaluation above, we retrained the best-performing model - random forest - using all the labelled datasets for each component in the pipeline. After retraining all models, we apply our workflow to the entire raw TED dataset. This contains roughly 3.3 million healthcare related tender notices (with contract awards) covering 2011 to 2022, involving over 167 thousands unique suppliers, 86 thousands buyers, with higher than $2 trillion in monetary value. Processing this massive dataset using our workflow explained above allowed us to create the biggest healthcare procurement database to date. We then run queries to obtain data from the database to calculate the above-mentioned metrics for each supplier. We show a few examples in screenshots below.

Figure 8 shows the supplier risk profile in terms of 'ability to supply' and 'economic risk' for Bausch & Lomb, based on their contracts won between 2011 and 2022. The line chart on the left shows a number of 'buyer' metrics (BM) selected for review, such as: 'buyer countries' that measures a supplier's global reach by considering countries they won contract in; 'buyers - moving average' that considers the number of active buyers for a supplier; and 'buyers - yearly participation' that considers the number of active buyers for supplier each year. The line chart on the right aggregates these selected metrics to show an overall trend. Figure 9 shows the supplier risk profile (also 'ability to supply' and 'economic risk') for Siemens covering the same time period, but using a mixture of 'lot' and 'buyer' metrics (LM and BM). For example, 'buyer - churn/retention rate' that measures the change in the supplier's clients (based on the number of new buyers they had and lost during each time period); 'lots - average duration days' and 'lots - duration days' looking at lot duration in days to understand lot delivery time frames. Each figure demonstrates risks of a specific supplier from different perspectives, hence allowing users to thoroughly evaluate a supplier.

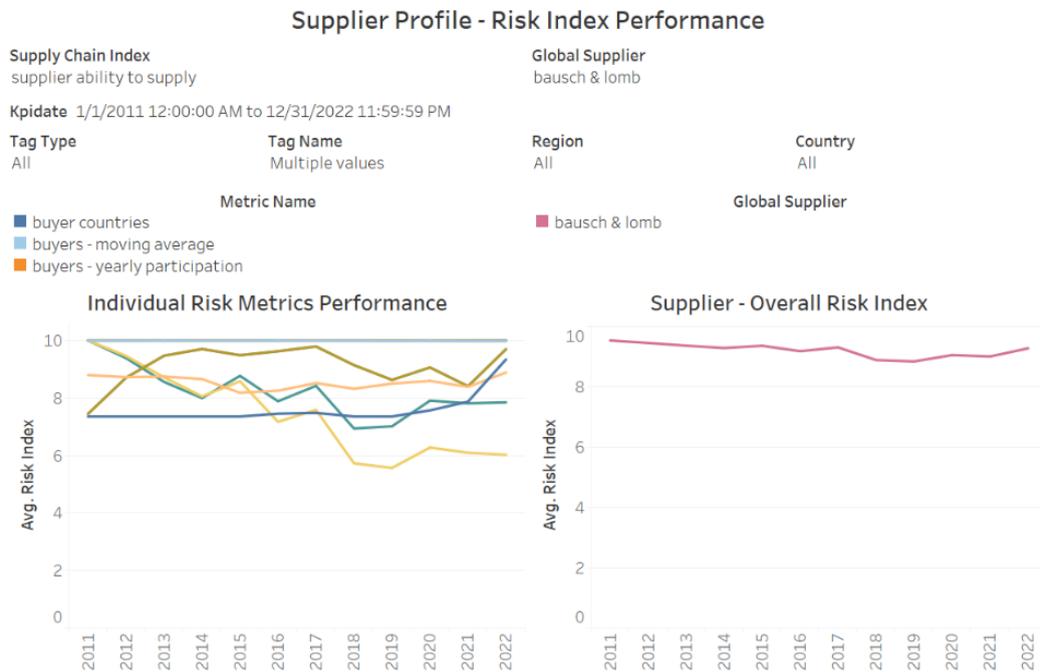

Figure 8. Supplier risk profile for Bausch & Lomb between 2011 and 2022 measured by a set of buyer metrics.

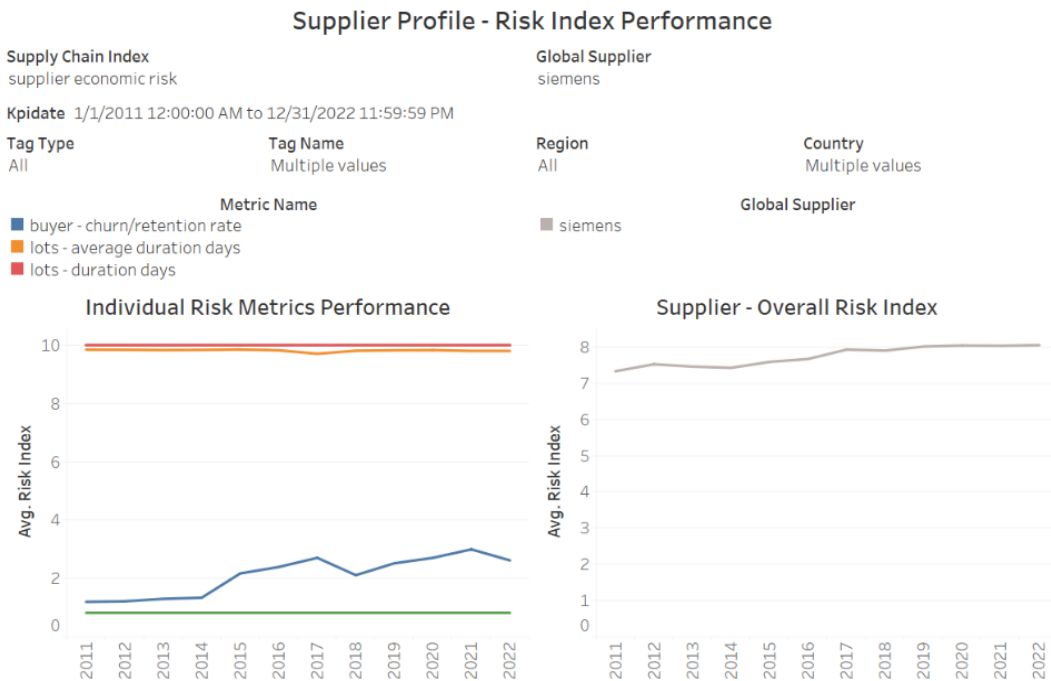

Figure 9. Supplier risk profile for Siemens between 2011 and 2022 measured by a set of lot metrics and buyer metrics.

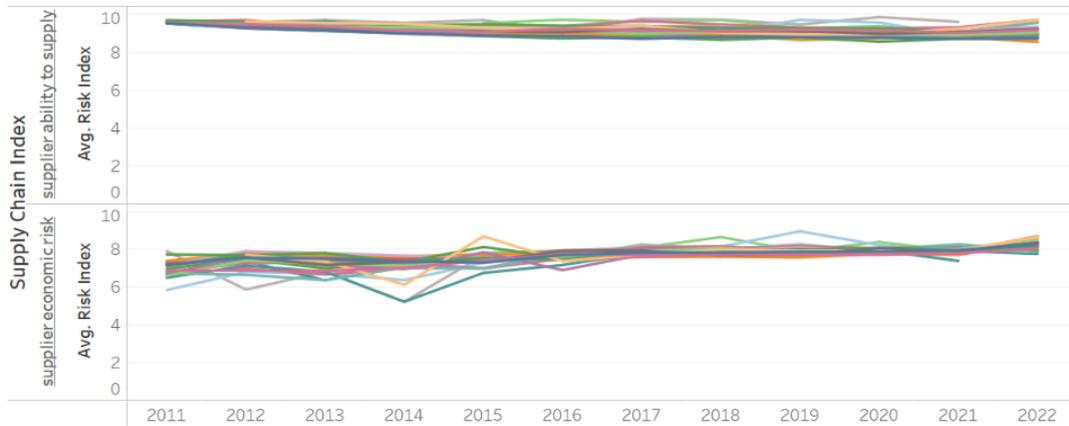

Figure 10. Comparison of several major global suppliers' risk profiles.

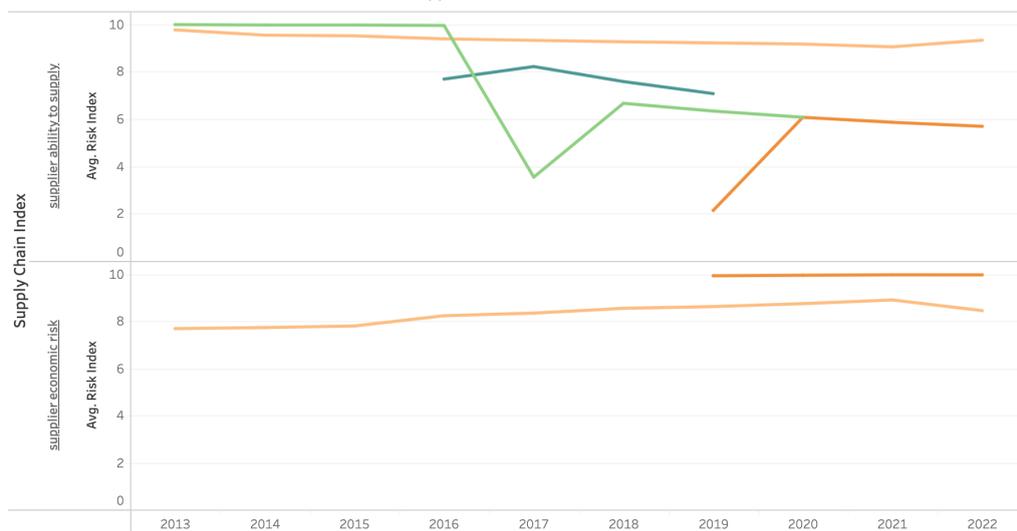

Figure 11. Comparison of several small suppliers' risk profiles.

Figure 10 and 11 compare global suppliers in a single view. Generally, a straight line with little fluctuations is desirable as that indicates little change in risks over time. We can notice that in Figure 10, most suppliers selected for review have relatively little change in terms of their risks. This is mainly due to them being large, established suppliers that tend to win a continuous stream of contracts over time. However, some suppliers had more fluctuations in their risk indices compared to others, suggesting they may be riskier choices to buyers. Figure 11 compares several smaller suppliers, and we can see that their risk patterns are much more erratic, due to a lack of continuity in their track record.

## 6. Discussion

In this section, we discuss lessons learned from the project that may potentially inform future research and practice. These will be covered from several angles: the different focus of industry project compared to research, the complexity of building a full NLP pipeline for heterogeneous data and its implication on the development, the dilemma of choosing between the more advanced NLP methods and those earlier classic methods, and the issue of training data in an industry context.

### *6.1. The 'industry' focus of the project*
There are very few studies that explicitly discuss the different focuses between industry and research projects and how these could impact the approach. Among these (Chiticariu et al., 2013; Suganthan et al., 2015; Krishna et al., 2016), it is widely acknowledged that the lack of training data and the associated cost of creating it, the fast development cycle, the need for interpretability of machine learning predictions, and the continuous update of the model due to evolving business needs and knowledge are factors that typically render a research-focused approach impractical. Industry projects focus on 'problem solving' in real world contexts with (often harsh) time and resource constraints, while research projects focus on 'creative solutions' to problems often studied in an ideal 'lab' environment that may not fully represent the reality.

Our experience has supported most - if not all - of the viewpoints above. The industry project has a clear focus of developing a 'proof-of-concept' product within a time limit of one year and financial budget. The business partner needs to be able to take the finished product for further development and/or extension beyond the project. Analysis at the beginning of the project showed its 'multi-task' and 'heterogeneous data' nature. Combining all these factors, a typical research-driven approach aimed at developing 'novel' solutions built on insights from rigorous critical evaluation of multiple baselines and state-of-the-art using research 'benchmarks' that may deviate from the real data, is infeasible. Instead, one lesson we learned is that one needs to opt for 'tried and tested' methods that have low 'barriers' to users, who may need to take the solution forward for future development. This is a fundamental principle that needs to be taken into account when evaluating other aspects of the project.

### *6.2. Data heterogeneity, multilingual and multi-task nature*
As already mentioned many times, compared to research as well as other industry text mining and NLP practices reported in the literature, the data we dealt with are multilingual and heterogeneous such that the goal of the project cannot be achieved by a single type of method, but relates to multiple subfields of text mining and NLP research. While there may be well-established solutions in each of these subfields, as discussed before, the difference in the raw data analysed, the multilingual nature and the lack of training data in our project makes adopting these methods very difficult and time-consuming. Considering the wider principle mentioned above, another lesson we learned is that given such a high level of complexity in our tasks, it is imperative to develop 'lightweight' methods that are easy to maintain, or 'generalised' solutions that could apply to multiple tasks, or both.

On reflection, we opted for treating many tasks as text classification, using a feature representation scheme that is arguably language independent, and applies to, or can be easily adapted to, all of the tasks. Specifically, two components of our pipeline (lot zoning and lot item detection) deal with text classification at different granularity and therefore, the machine learning algorithms can be easily reused across. The feature representations are derived with reference to domain lexicons, which can be more easily and affordably translated into multiple languages - particularly considering the massive amount of multilingual documents we have to analyse, and the fact that only a fraction of them contain really useful content for analysis. Another added benefit of our approach is that it enables domain experts and the system administrators to update the trained models by simply revising the domain lexicons during re-training. To the best of our knowledge, there is no prior literature that describes in detail a holistic text mining or NLP process composed of multiple sub-tasks, or develops generalisable solutions like ours. Therefore, our work sets an important reference for future industry projects dealing with complex tasks.

### *6.3. The dilemma of algorithmic choices*
Our classification method is based on 'classic' machine learning algorithms and does not utilise the more recent deep neural networks trained on very large datasets as 'language models', such as the well known

BERT (Delvin et al., 2018). These models have been shown to set new benchmarks for a wide range of NLP tasks. However, we did not use BERT or similar language model based methods for a number of practical reasons.

First, such models work in slightly different ways from classic algorithms in the sense that they take over the feature extraction process through a 'text reading' process analogous to humans. The idea is that through training such models using enormous data, they are expected to have a certain level of 'understanding' of human language, and therefore, can learn to automatically extract useful features from the input texts. This makes feature engineering not very compatible with such models, although one can still run such models as a 'feature extractor' to generate a feature representation vector, and combine it with other predefined features.

Second, such models are usually very resource intensive due to its complex architecture and the amount of data used to train them. As a result, they usually can only take a limited length of text. Popular options are a length of 256 or 512 tokens, with the second typically requiring significantly higher computing resources. This means that some of our tasks, such as page or table classification, may not fit well when the content is longer than this length, particularly the smaller option which is more affordable.

Third, we experimented with a generic BERT model (English uncased, 256) on the lot item detection task using English training data only, and compared its performance against our best performing model, random forest. However, we did not notice significant gain in F1 by BERT (under 2 percentage points). We assume that a potential reason is that the generic BERT model is trained on general purpose corpus that may not be closely related to healthcare. Studies (Alsentzer et al., 2019; Beltagy et al., 2019) have shown that usually, language models need to be further trained using domain specific corpora when they are to be applied to domain specific contexts. This again requires significant data and computing resources, especially when we have to deal with multiple languages.

Therefore, the next lesson we learned is that for industry projects, practicality is an important factor when it comes to choosing the methods. Resource constraints often outweigh algorithmic superiority, especially if the performance gain is small given the potential resource investment needed.

### 6.4. The cost of training data

In the project, we used a mixture of supervised and unsupervised (rules) methods. There are many practical reasons for this choice, but contrary to the scientific literature that is predominantly based on supervised machine learning methods (Suganthan et al., 2015), the main reasons for not opting for a fully supervised approach is the cost. This may be explained from many factors, such as the human labour, the time required, the number (and complexity) of tasks to be addressed, and the tools needed for developing the training data.

The project uses a process that involves multiple tasks of varying complexity. Developing training data for the two zoning tasks is relatively low-cost, as they do not require specialist tools to support the annotation process. In both tasks, annotators were given the extracted text passages either in raw text or CSV (for tables only), together with the original document for reference. The annotation task is also fairly easy, as it only requires skimming through these documents, and does not require extensive reading. In contrast, annotating lot items (lot item detection) and their specific attributes (lot parsing) are much lower granularity tasks, and require examining content at sentence or phrase level. The data also needs to be transformed into formats that can be loaded into state of the art annotation tools that are not always available for our data format. For example, the widely used 'spacyNER' annotation format requires parallel sequences of tokens and their labels for each sentence. Such data is very expensive to develop from a business perspective.

Further, as already raised in earlier studies, industrial solutions often favour rules over supervised models as the latter offers easier interpretability and maintainability. These are crucial factors as the needs for their applications can evolve over time and therefore, the solutions must be easy to modify over time. In these cases, changing rules may be a much lighter effort than recreating training data and retraining the system. For these reasons, training data for supervised models can be a barrier to the adoption of methods developed in the research communities, when it comes to developing industrial applications.

### 7. Conclusion

Text mining and NLP have been long established research fields for decades. Their techniques have also been widely adopted in industries to develop and deploy intelligent systems for automated analysis of large scale text data. However, the literature is dominated by research that favours supervised methods built on well-curated data. Solutions in such a 'lab environment' often do not transfer well to practical scenarios. Instead, studies reporting industrial text mining/NLP tasks often make use of rule-based methods and domain lexicons. But they typically look at single and sometimes simplified tasks, and do not discuss in-depth data heterogeneity and inconsistency and their implication on the development of their methods. Further, few prior work has focused on the healthcare domain.

Set in this context, our work describes an industry project that developed text mining methods and solutions to mine millions of heterogeneous, multilingual procurement documents in the healthcare sector. We extract structured procurement contract data and store them in a database that drives a platform enabling easy evaluation of supplier risks. Our work sets reference for future research and practice in many ways: 1) it develops the first structured procurement contract database that will help facilitate the tendering process; 2) it documents a method that effectively uses domain knowledge and generalises to multiple text mining and NLP tasks and languages; 3) and it discusses lessons learned for practical text mining/NLP development.

Drawing from our lessons, we make a few recommendations for researchers and practitioners. First, we argue that research needs to 'step out of the lab environment' by using data that more reflects reality. Research data are typically well-curated and pre-processed. But as we have seen, in practice, real data is rarely good quality and highly inconsistent. This means that practitioners often need to make a significant effort to cleanse their data, or adapt state-of-the-art from research. Both are non-trivial. Also, rules continue to be important and effective in many real world applications, as they are easier to implement, fit for purpose, and easy to interpret. We believe an interesting direction is for model explainability research to develop methods that can explain model decisions in terms of rules beyond the current 'primitive' approaches (e.g., feature weights, attentions). These may offer valuable insights for building domain-specific applications. Third, for practitioners, we recommend that they focus on their real needs when it comes to algorithmic choices. While the recent text mining and NLP research has seen deep neural networks - especially very large language models trained on massive corpora - taking over the centre stage, the added value to businesses in practice may depend on the domain and task. This is particularly important if the business has restricted access to resources, as these methods are much more resource intensive than classic machine learning models. Finally, building industrial text mining and NLP applications usually entails a process involving multiple tasks. While often, there can be tried-and-tested methods for each task, one needs to again consider their resource constraints and it helps to think in terms of building solutions that can generalise to a wide range of tasks instead of buying or adapting ad-hoc solutions for each task.

Our work has a number of limitations. First, we have not evaluated the end system, i.e., the platform for deriving supplier risk profiles. This is primarily due to the work being taken further for development by the industry partner before being presented to end users. An end-user evaluation would be an extremely valuable exercise to examine the effectiveness of our text mining and NLP methods. Second, our work has focused on a specific sub-area of healthcare - pharmaceuticals. This is arguably an easier sub-area compared to medical equipment where the naming and standards can be very inconsistent. Therefore, it is difficult to conclude how our methods can generalise to these areas.

In terms of future work, we identify three main directions. First, we will look at adapting our solution to other areas of the healthcare sector (e.g., medical equipment as mentioned above), or other sectors. Second, while within the project, we only analysed procurement documents, another source of useful information is supplier websites and their product catalogues. We envisage to mine such data in the future to enrich our database. Finally, we recognise a lack of research in the area of procurement text mining and NLP. For this reason, we plan to release part of our data (subject to further processing to redact sensitive information) for use by the research community and set up shared task to encourage effort on this direction.

## Acknowledgements

Part of this work was funded by the Innovate UK under the project 90205 'AI-powered real-time healthcare supplier profile and COVID-19 supply risk matrix'.